\documentclass[runningheads]{llncs}
\newcommand{\upcite}[1]{\textsuperscript{\textsuperscript{\cite{#1}}}}
\usepackage{graphicx}
\usepackage{CJKutf8}
\usepackage{listings}
\usepackage{booktabs}
\usepackage{algorithm}  
\usepackage{algpseudocode}  
\usepackage{amsmath}  
\usepackage{marvosym}

\begin{document}
\begin{CJK}{UTF8}{gbsn}
    \title{Bilevel Scheduled Sampling for Dialogue Generation
    }

    \author{Jiawen Liu
        \and
        Kan Li\textsuperscript{(\Letter)}
    }
    \authorrunning{J. Liu and K. Li}

    \institute{
        Beijing Institute of Technology, Beijing, China\\
        \email{\{liujiawen,likan\}@bit.edu.cn}}

    \maketitle              
    \begin{abstract}

        Exposure bias poses a common challenge in numerous natural language processing tasks, particularly in the dialog generation.
        In response to this issue, researchers have devised various techniques, among which scheduled sampling has proven to be an
        effective method for mitigating exposure bias.
        However, the existing state-of-the-art scheduled sampling methods solely consider
        the current sampling words' quality for threshold truncation sampling,
        which overlooks the importance of sentence-level information and the method of threshold truncation warrants further discussion.
        In this paper, we propose a bilevel scheduled sampling model that takes the sentence-level
        information into account and incorporates it with word-level quality.
        To enhance sampling diversity and improve the model's adaptability,
        we propose a smooth function that maps the combined result of sentence-level
        and word-level information to an appropriate range,
        and  employ probabilistic sampling based on the mapped values instead of threshold truncation.
        Experiments conducted on the DailyDialog and PersonaChat datasets
        demonstrate the effectiveness of our proposed methods,
        which significantly alleviate the exposure bias problem and outperform state-of-the-art scheduled sampling methods.

    \end{abstract}
    \section{Introduction}
    Exposure bias is a common problem in many natural language processing tasks \cite{2015Scheduled},
    especially in dialog generation.
    Exposure bias refers to the discrepancy between the training and inference stages of a model,
    where the model is trained on ground truth data but generates on its own predictions at inference time.
    This results in the model facing different environments during training and inference.
    Once a model prediction is inconsistent with the ground truth somewhere,
    it can affect later predictions, which will lead to errors accumulating and propagating along the generated sequence,
    resulting in poor quality, diversity and generalization
    of responses.

    To address this issue, researchers have developed various techniques,
    such as beam search \cite{beamsearch},
    DAD(Data As Demonstrator) \cite{2015Improving},
    sentence level training \cite{2015Sequence}, scheduled sampling \cite{2015Scheduled,2019Scheduled},
    reinforcement learning explicitly trains models \cite{2016SeqGAN,2018RelGAN}, etc.
    These techniques help models handle long-term dependencies better in inference process.
    Currently, scheduled sampling is widely adopted and has demonstrated favorable outcomes.
    Nevertheless, the existing methods either generate entire
    sentences autoregressively before sampling\cite{zhang-etal-2019-bridging},
    or solely consider the current sampling word's impact
    for threshold truncation sampling\cite{2021Adaptive,2021Confidence}.
    The former necessitates multiple beam searches,
    resulting in high computational complexity, low efficiency,
    and cannot dynamically adjust based on the quality of each word,
    thereby overlooking the differences between words in the sentence.
    The latter disregards sentence-level information and
    the method of threshold truncation warrants further discussion.

    In our opinion, the importance of word-level information is obvious,
    and current methods that sample based on word-level scores \cite{2021Adaptive,2021Confidence}
    are significantly better than traditional methods \cite{2015Scheduled,zhang-etal-2019-bridging}
    that sample each word in a sentence with the same probability.
    However, relying on cosine similarity or other scores
    between the generated word and the ground truth word is limited,
    and we still need to take the sentence-level quality into account.
    ​Therefore, in this paper, we propose a bilevel scheduled sampling model that
    dynamically adjusts the sampling probabilities of the generated results
    and the ground truth at both word and sentence level.​
    Our model effectively balances the trade-off between learning from the ground truth
    and its own predictions, thereby enhancing the robustness and diversity of dialogue generation.



    ​Specifically, sentence quality evaluation in sentence-level sampling is conducted
    through the utilization of either BLEU score or sentence-level cosine similarity.
    On the other hand, word quality in word-level sampling is evaluated based on the probability of word generation.
    Finally, a specially designed smooth function is used to
    integrate sentence-level and word-level evaluations, enabling probabilistic sampling.​

    ​We evaluate the proposed model on two publicly available datasets,
    namely DailyDialog and PersonaChat.
    Experimental results show that our model is superior to
    existing  exposure bias models
    in both metric-based automatic evaluation and human evaluation.
    The main contributions of this paper include:
    \begin{itemize}
        \item ​To the best of our knowledge, we are the first to propose
              a bilevel scheduled sampling model that takes both sentence-level
              and word-level information into account.
        \item In sentence-level sampling, we utilize BLEU and sentence-level cosine similarity as evaluation metrics.
              For word-level sampling, we leverage the predicted probabilities as the word-level score.
              To enhance the diversity of the sampling process and improve the model's adaptability,
              we propose a smoothing function that maps the combined results of bilevel sampling to an appropriate range
              and adopt probabilistic sampling instead of threshold truncation.
        \item Extensive evaluations on two widely used open-domain dialogue datasets
              demonstrate that the proposed approach significantly alleviates
              the exposure bias problem and outperforms the state-of-the-art scheduled sampling methods.
    \end{itemize}

    \section{Related Work}
    Data As Demonstrator (DAD) is a meta learning algorithm \cite{2015Improving}
    that solves the problem of exposure bias
    by blending real training data with model predictions.
    During the training phase, not only the ground truth is used as input,
    but also the model-predicted results are added to
    the training set to make it conform to the test distribution.
    The Scheduled Sampling method \cite{2015Scheduled} further developed this approach for sequence generation
    by sampling from the model's own predictions during training, instead of always using the ground truth.
    This method was first proposed by Bengio et al. in 2015 \cite{2015Scheduled},
    and was later improved for transformer models by Mihaylova and Martins in 2019 \cite{2019Scheduled},
    owing to the superior performance of transformer models \cite{2017Attention}.
    What's more, since scheduled sampling only looks ahead one step, Goodman et al. proposed the TeaForN method \cite{2020TeaForN},
    which looks ahead N steps for more foresight in sampling,
    but at the cost of reduced training efficiency.

    The initial strategy for sampling was to decay the probability
    of sampling the ground truth based on training  steps,
    with specific linear decay $f(i)=max(\epsilon,ki+b)$,
    exponential decay $f(i)=k^i$, and sigmoidal decay $f(i)=\frac{k}{k+e^{\frac{i}{k}}}$.
    Later in 2021, Liu et al. proposed a strategy of decaying according to decoding
    steps \cite{2021Scheduled}, further improving the effectiveness.

    Zhang et al. proposed two strategies,
    namely word-level and sentence-level oracles,
    to select the generated results of the model \cite{zhang-etal-2019-bridging}.
    However, the article continues to employ probability with decay for sampling
    from model's results (oracles) and the ground truth, rather than adopting a more
    sophisticated approach of sampling based on the quality of sentences or words.

    Obviously, sampling words with equal probability regardless of their varying qualities is inherently inaccurate.
    It is highly recommended to employ distinct sampling probabilities for individual sentences and for each word within a sentence.
    Liu et al. proposed confidence-aware scheduled sampling \cite{2021Confidence},
    which sets two thresholds $t_{golden}$ and $t_{rand}$, and samples the ground truth, model prediction result,
    a random word when the confidence is respectively in  $[0,t_{golden})$, $[t_{golden}, t_{rand})$,  $[t_{rand}, 1]$.
    Xu et al. \cite{2021Adaptive} focused on dialogue generation
    and selected words based on the cosine similarity between
    the predicted word and the ground truth.
    If the similarity is greater than the threshold $\beta$,
    the predicted word is selected with a probability of $\alpha$,
    which is a hyperparameter that increases with the number of training
    epochs to achieve faster convergence in the early stages of
    training and alleviate exposure bias.

    However, these methods have some limitations.
    Firstly, they don't take into account the very important sentence-level information.
    Secondly, 
    the method of threshold truncation also needs to be discussed.
    If probability selection is used, different words can be sampled with different probabilities,
    so that the distinction between different words is greater and the adaptability of the model is stronger.


    Consequently,  we propose a novel bilevel scheduled sampling model that effectively
    combines sentence-level information with word-level quality.
    To enhance sampling diversity and improve the model's adaptability,
    we introduce a smooth function that maps the integrated outcome of sentence-level and word-level information to a suitable range.
    Subsequently, we employ probabilistic sampling based on the mapped values instead of threshold truncation.

    \section{Bilevel Scheduled Sampling Model}

    \subsection{Mathematical Modeling of Dialogue Generation}
    In dialogue generation tasks, we typically define the model's objective function
    by maximizing the conditional probability, which means we need to maximize
    the probability of generating the output text $Y$ given the input text $X$.
    Specifically, we can use equation \ref{equation1} to represent this objective function,
    where $y_t$ is the $t$-th word in $Y$, $y_{<t}$ are the first $t-$1 words in $Y$,
    and $T$ is the length of  $Y$. This objective function
    requires us to calculate the probability of each word given the input text $X$
    and the previous words $y_{<t}$,
    and multiply these probabilities to obtain the probability of the output text $Y$.

    \begin{equation}\label{equation1}
        P(Y|X,\theta) = \prod_{t=1}^{T}p(y_t|y_{<t},X,\theta)
    \end{equation}

    During the training process, we typically use
    the cross-entropy loss function \cite{CrossEntropy}
    as the optimization objective for the model,
    which measures the difference between the model's output
    and the ground truth. Specifically, we can use equation \ref{equation2}
    to represent the cross-entropy loss function, where $L(\theta)$
    denotes the cross-entropy loss function of the model,
    and the negative logarithm of the probability is used as the loss,
    such that the loss decreases as the probability increases.

    \begin{equation}\label{equation2}
        L(\theta)=- \log P\left(Y\middle| X,\theta\right)=\frac{1}{T}\sum_{t=1}^{T}{- \log p\left(y_t\middle| y_{<t},X,\theta\right)}
    \end{equation}

    By minimizing the loss function of all samples, we can obtain the optimal model parameters $\theta$.
    This can be expressed as equation \ref{equation3}, where $N$ is the total number of training set samples.

    \begin{equation}\label{equation3}
        \theta = \mathop{\mathrm{argmin}}\limits_{\theta} \{\sum_{k=1}^{N}L_k(\theta)\}
    \end{equation}

    In exposure bias problem, unlike training, during inference,
    the probability of each target word $p(y_t|y_{<t},X,\theta)$ in equation \ref{equation1}
    is conditioned on the previously generated words $y^*{<t}$ instead of
    the ground truth $y{<t}$, because the ground truth words are not available in actual inference.
    Therefore, we employ the scheduled sampling method to replace the ground truth during training
    with sampled sentences, thereby reducing the gap between the model's training and inference processes.

    \subsection{Bilevel Scheduled Sampling}
\begin{figure}
        \vspace{-0.1cm}
        \centering
        \includegraphics[width=1\textwidth]{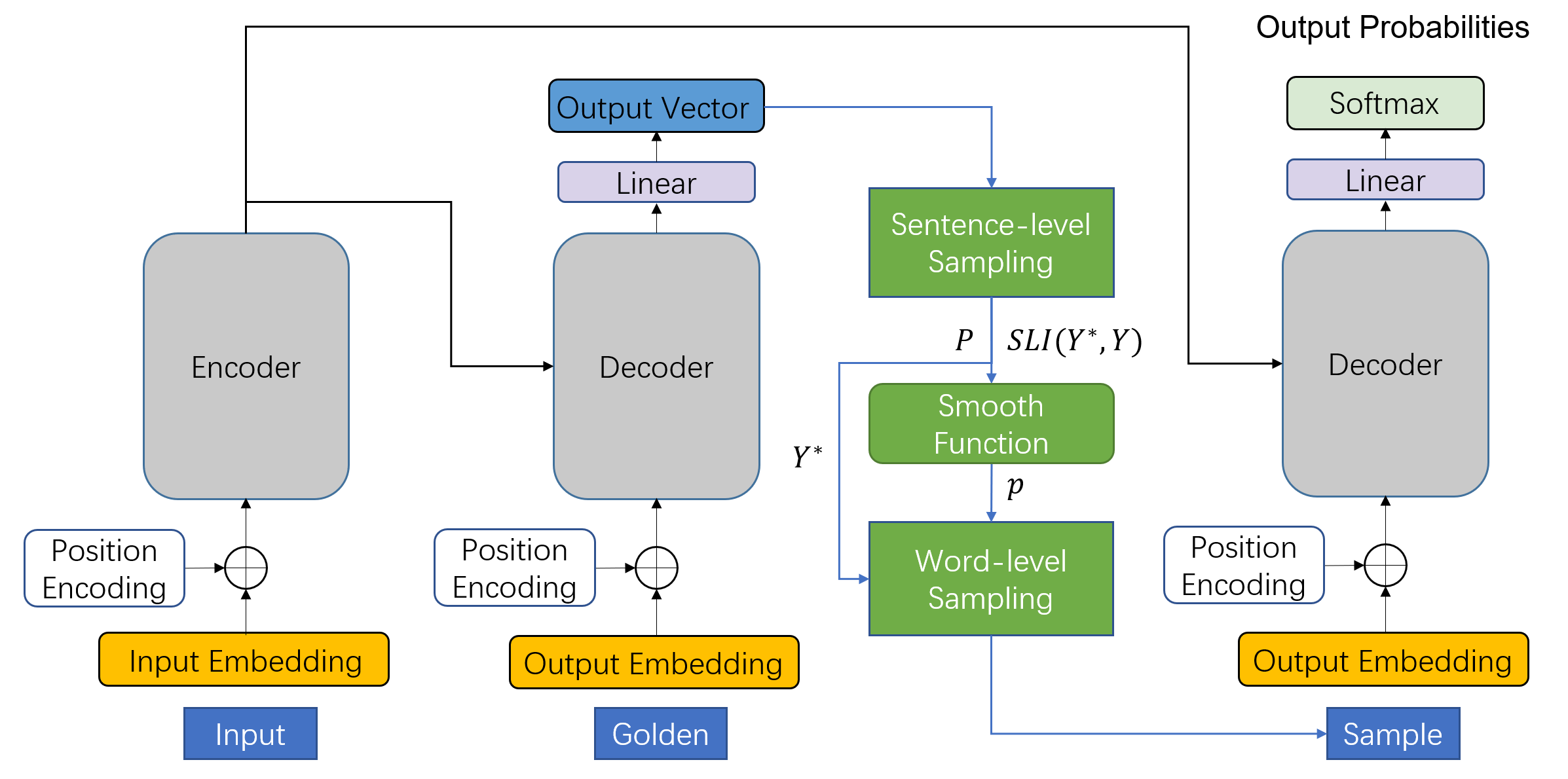}
        \caption{Our Bilevel Scheduled Sampling model.} \label{model}
        \vspace{-0.2cm}
    \end{figure}

    We propose a sentence- and word-level fusion
    sampling method to address the diversity and contextuality
    of open-domain dialogue systems.
    Taking the transformer as an example,
    we introduce the principle and implementation
    of the bilevel scheduled sampling method, as shown in Fig. \ref{model}.

    \subsubsection{Sentence-level Sampling}

    Zhang et al. \cite{zhang-etal-2019-bridging} proposed a traditional sentence-level sampling method
    that uses beam search to select k most likely sentences from the predicted distribution,
    and then chooses one as the sampling sample based on the BLEU score,
    followed by the decay sampling method to extract from the sample and the real sentence.
    However, this method has two main drawbacks:
    (1) it requires multiple beam searches to generate sentence
    like inference procedure, resulting in high computational cost and low efficiency;
    (2) it can only sampling  by probability with decay,
    which ignores the differences between words in the sentence,
    making it impossible to dynamically adjust the quality of each word.

    To address these issues, we propose a sentence-level sampling method
    that utilizes the parallelism of the transformer model to generate the entire
    sentence at once during training, and calculates the probability of each word
    and the quality of the sentence.

    We first calculate the probability $P$ of the predicted word through the softmax function,
    and select the words $Y^*$ with the largest probability to be evaluated by the sentence-level indicator.

    We try two sentence-level indicators(SLI):
    BLEU and sentence-level cosine similarity, both of which have achieved good results.
    And considering the padding tokens in the sentences, we mask them during calculation.
    One metric based on BLEU is as follows:
    \begin{equation}
        \operatorname{SLI1}(Y^*,\, Y) = \frac{1}{m}{\sum_{i=1}^{4}\operatorname{bleu-i}(Y^*,\, Y)}
    \end{equation}
    Where the term `$\operatorname{bleu-i}$' refers to the i-gram BLEU result
    without using a smooth function \cite{papineni2002bleu}.
    $Y^*$ represents the model-generated result,
    $Y$ represents the ground truth,
    and $m$ is a hyperparameter used to map the result to a value around 1 to prevent imbalance during sampling.

    The other metric based on cosine similarity is as follows:
    \begin{equation}
        \begin{aligned}
            \operatorname{SLI2}(Y^*,\, Y) & = \frac{1}{m} \operatorname{CosinSimilarity}(Y^*, \, Y)                                                                                    \\
                                          & =  \frac{\operatorname{embed}(Y^*) \cdot \operatorname{embed}(Y)}{m \cdot ||\operatorname{embed}(Y^*)|| \cdot ||\operatorname{embed}(Y)||}
        \end{aligned}
    \end{equation}

    Where $m$ is the hyperparameter, which maps the result to a value around 1 so that the sampling is not unbalanced;
    $\operatorname{embed}$ is word embedding and converts word subscript to word vector.
    We just use word embedding within the model's decoder. An existing word embedding
    from pretrained model such as BERT\cite{BERT} can also be used.
    Here, for sentences, it is the average word embedding of sentences:
    \begin{equation}
        \operatorname{embed}(Y) = \frac{\sum_{i=1}^{T}\operatorname{embed}(y_i)}{T}
    \end{equation}
    Where $T$ is the length of sentence $Y$, $y_i$ is the i-th word.

    This method has the following advantages:
    First, the whole sentence can be generated at once, avoiding the cost of multiple beam searches and improving efficiency.
    Second, it considers the sentence-level information and can be combined with the word-level sampling to dynamically adjust each word,
    enables the model to sample based on the quality of individual words
    while simultaneously accounting for the overall quality of the sentence.

    \begin{figure}[htbp]
        \begin{minipage}[t]{0.5\linewidth}
            \centering
            \includegraphics[width=0.7\textwidth]{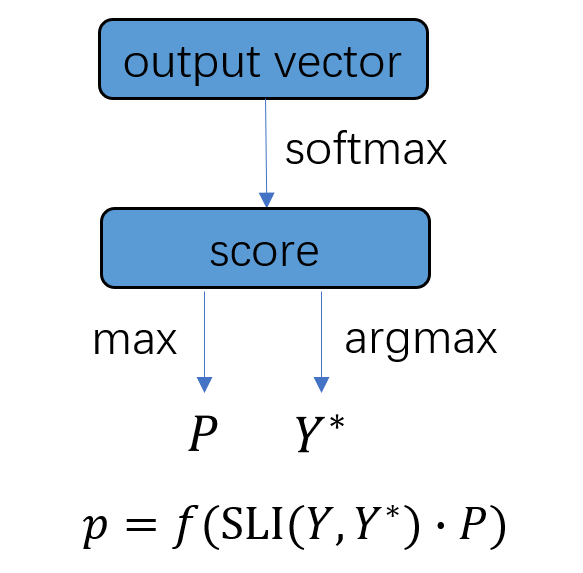}
            \caption{Sentence-level Sampling}
            \label{fig:sentence sampling}
        \end{minipage}%
        \begin{minipage}[t]{0.5\linewidth}
            \centering
            \includegraphics[width=0.7\textwidth]{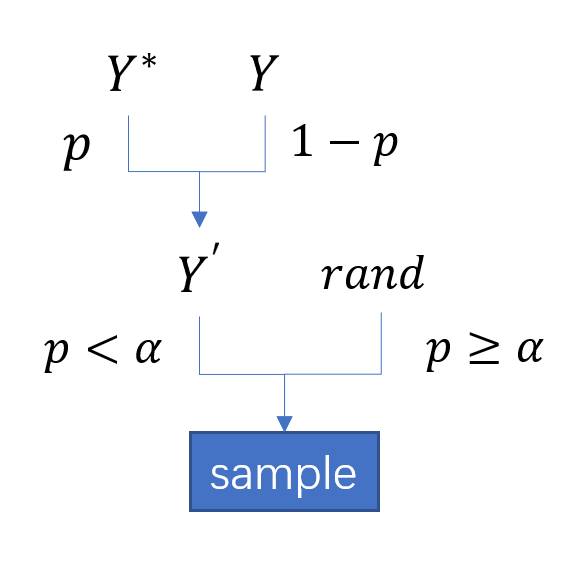}
            \caption{Word-level Sampling}
            \label{fig:word sampling}
        \end{minipage}
    \end{figure}
    \subsubsection{Smooth Function Between Two Sampling Layers}
    In order to simulate the inference process,
    words with higher corresponding probabilities $P$
    have a higher probability of being sampled,
    so we directly use $P$ as the word-level evaluation metric.
    Now we have the sentence-level score $S$(result of $SLI$) and the word-level score $P$.
    However, a method to combine them is still missing.
    Considering that both sentence-level score $S$ and
    word-level score $P$ are equally important,
    we multiply them and then use a smooth function
    to map the product to the range of 0\textasciitilde 1, like Fig. \ref{fig:sentence sampling}.

    Currently,  we take two types of functions into account.
    The first one is simpler, which directly restricts the result to 0\textasciitilde 1, like equation \ref{equation5}.
    \begin{equation} \label{equation5}
        f(x) = max(min(x,\, 1),\, 0)
    \end{equation}

\begin{figure}
        \centering
        \setlength{\belowcaptionskip}{-0.9cm}
        \includegraphics[width=0.9\textwidth]{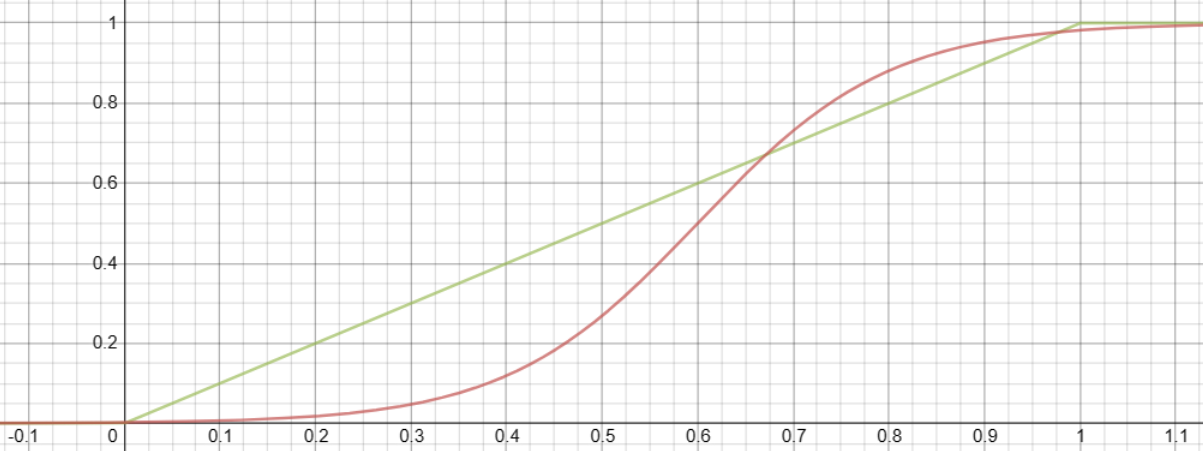}
        \caption{The smooth function $f$.} \label{f}
    \end{figure}

    The second one is a sigmoid-shaped function which is smoother, like equation \ref{equation6}.
    \begin{equation} \label{equation6}
        f(x) = \frac{1}{1+e^{-k\left(x-b\right)}}
    \end{equation}
    where $k \geq  1$ is a hyperparameter to control the speed of convergence,
    $b > 0 $ controls the central symmetrical point of the function, moving it to the right.
    We finally set $k=10,b=0.6$. Their images are in Fig. \ref{f}.

    \subsubsection{Word-level Sampling}
    Now, we have obtained the smoothed score $p$,
    which contains information combining sentence-level and
    word-level, and we use it for specific word sampling as shown in Fig. \ref{fig:word sampling}.

    First, we use probability selection for sampling like equation \ref{equation9},
    which has a greater discrimination between different words
    compared to threshold truncation.
    It can also sample different words,
    making the trained model more adaptable.
    Specifically, the predicted word $Y^*$ is selected with probability $p$ and
    the ground truth $Y$ is selected with probability $1-p$, and $Y^{'}$
    is the first sampled result.
    \begin{equation} \label{equation9}
        Y^{'}  = \begin{cases}
            Y^* & sampling \enspace with \enspace probability \enspace p \, , \\
            Y   & otherwise.
        \end{cases}
    \end{equation}

    In addition, to prevent the model from relying too heavily
    on high-probability predicted words and
    causing the generated results to be too monotonous,
    this paper also uses a method for
    random word sampling, like equation \ref{equation10}.
    When the word-level predicted probability $P$
    is greater than a set threshold $\alpha$,
    a random word is chosen as the next input
    with a certain probability to prevent
    the model from degenerating.
    \begin{equation} \label{equation10}
        sample = \begin{cases}
            Y^{'} & if \enspace P<\alpha \, , \\
            rand  & otherwise.
        \end{cases}
    \end{equation}
    where $rand$ refers to a random word,
    $\alpha$ is a threshold which is set to 0.95,
    $sample$ is the finally sampled result as the input to encoder during training.

    \section{Experiments}


    To assess the effectiveness and merits of the proposed Bilevel Scheduled Sampling model
    detailed in this study, a comprehensive set of experiments was conducted,
    encompassing evaluation, comparative analysis and ablation study.
    This section elucidates the experimental design employed and
    provides a thorough evaluation of the obtained results.

    \subsection{Datasets}
    We evaluate the proposed method using two widely used dialogue datasets.
    DailyDialog is a collection of daily life conversations, encompassing
    a variety of topics, emotions and linguistic styles\cite{dailydialog}.
    PersonaChat consists of conversations between two participants,
    where one participant assumes a persona and the other participant engages
    in a conversation while considering the persona's characteristics\cite{personachat}.
    After data preprocessing, we split the n-turn dialogue
    $(u_1,u_2,...,u_n)$
    into n-1 single-turn dialogues
    $[(u_1,u_2),(u_2,u_3),...,(u_{n-1},u_n)]$
    , where $u$ represents an utterance.
    The number of context-response pairs in the train/validation/test set is 68,066/6,820/6,841
    for DailyDialog and 104,609/12,588/12,106 for PersonaChat
    without any extra label or persona information.

    \subsection{Implementation Details}
    The experiment uses an NVIDIA GeForce RTX 2080 Ti graphics card and adopts PyTorch deep learning framework for training.
    Dropout is used for the selfattention module, the feed-forward layer, and the activation layer, and the rate of all
    three is set to 0.1. The sentence length is set to 26 and batch size is set to 256.
    The vocab size is 21626 for DialyDialog and 22630 for PersonaChat.

    \subsection{Comparison Methods}
    We compare our proposed Bilevel Scheduled Sampling model with
    following established methods, and all approaches are based
    on the Transformer-base model\cite{2017Attention}:
    \begin{itemize}
        \item \textbf{Transformer}\cite{2017Attention}: The Transformer-base model used in dialog generation.
        \item \textbf{AdapBridge}\cite{2021Adaptive}: An improved scheduled sampling approach, which uses an adaptive bridge mechanism
              to evaluate model generation results. Specifically, the selection is made
              according to the cosine similarity results of the predicted word and the ground truth.
              If it is greater than a threshold $\beta$, the predict word is sampled according to
              a probability $\alpha$ that varies with the number of training epochs.
              Accoding to the paper, we set $w=15$ as half of the training epochs and other hyperparameters are not changed.
        \item \textbf{Confidence-Aware}\cite{2021Confidence}: An improved scheduled sampling method, which selects
              whether to sample according to the confidence of the prediction result (that is, the prediction probability).
              Specifically, it sets two thresholds $t_{golden}$ and $t_{rand}$, and samples the ground truth, model prediction result,
              a random word when the confidence is respectively in  $[0,t_{golden})$, $[t_{golden}, t_{rand})$,  $[t_{rand}, 1]$.
              We set $t_{golden}=0.7$ and $t_{rand}=0.95$  to get the best result.
    \end{itemize}
    At the same time, ablation experiments are carried out in this paper.
    For different models, we tested different hyperparameter values for $m$,
    and presented the best result.
    The proposed model for ablation experiment testing is as follows:

    \begin{itemize}
        \item \textbf{Bilevel-None}: The bilevel scheduled sampling model proposed in this paper without the sentence-level sampling part.
              The smooth function is the sigmoid-shaped equation \ref{equation6} and the word-level sampling method is unchanged.
        \item \textbf{Bilevel-Bleu}: The bilevel scheduled sampling model proposed in this paper, the  sentence-level indicator
              is the Bleu metric with $m=0.8$ to get the best result.
              The smooth function is equation \ref{equation6}.
        \item \textbf{Bilevel-Cosine}: The bilevel scheduled sampling model proposed in this paper, the  sentence-level indicator
              is the sentence-level cosine similarity with $m=0.6$ to get the best result.
              The smooth function is equation \ref{equation6}.
        \item \textbf{Bilevel-f1}: The bilevel scheduled sampling model proposed in this paper, the  sentence-level indicator
              is the Bleu metric with $m=0.9$ to get the best result.
              The smooth function is the linearly truncated equation \ref{equation5}.
    \end{itemize}

    \subsection{Automatic Evaluation}
    We evaluate the performance of dialogue generation where both automatic and human evaluation metrics are applied.
    Automatic evaluation metrics include BLEU-1/2/3/4 \cite{papineni2002bleu}, Distinct-1/2/3 \cite{li-etal-2016-diversity}.
    The result is shown in Table \ref{tab:result}.

    \begin{table*}[!ht]
        \centering
        \caption{Evaluation results on Daily Dialog and Persona Chat datasets.}

        \begin{tabular}{l|cccc|ccc}
            \toprule
            \multicolumn{8}{c}{Daily Dialog}                                                                                                                                                                    \\
            \toprule
            Model                                   & \multicolumn{4}{c|}{BLEU-1/2/3/4} & \multicolumn{3}{c}{Distinct-1/2/3 }                                                                                   \\
            \toprule
            Transformer\upcite{2017Attention}       & 16.47                             & 5.96                                & 3.30           & 2.11          & 0.90          & 4.53          & 11.34          \\
            AdapBridge\upcite{2021Adaptive}         & 16.78                             & 6.06                                & 3.51           & 2.18          & 0.85          & 4.24          & 10.13          \\
            Confidence-Aware\upcite{2021Confidence} & 16.63                             & 6.45                                & 3.65           & 2.25          & 0.89          & 4.40          & 10.73          \\
            Bilevel-None                            & 16.84                             & 6.53                                & 3.66           & 2.33          & \textbf{0.99} & 4.86          & 11.69          \\
            Bilevel-Bleu                            & \textbf{17.43}                    & \textbf{6.81}                       & \textbf{3.87}  & \textbf{2.49} & 0.98          & \textbf{5.10} & \textbf{12.98} \\
            Bilevel-Cosine                          & 17.24                             & 6.73                                & 3.79           & 2.45          & 0.94          & 4.65          & 11.43          \\
            \toprule
            \multicolumn{8}{c}{Persona Chat}                                                                                                                                                                    \\
            \toprule
            Model                                   & \multicolumn{4}{c|}{BLEU-1/2/3/4} & \multicolumn{3}{c}{Distinct-1/2/3 }                                                                                   \\
            \toprule
            Transformer\upcite{2017Attention}       & 17.79                             & 6.37                                & 3.42           & 2.31          & 0.22          & 0.65          & 1.34           \\
            AdapBridge\upcite{2021Adaptive}         & 19.53                             & 6.79                                & 3.65           & 2.46          & 0.20          & 0.58          & 1.20           \\
            Confidence-Aware\upcite{2021Confidence} & 20.15                             & 7.35                                & 3.82           & 2.48          & 0.19          & 0.62          & 1.36           \\
            Bilevel-None                            & 19.84                             & 7.20                                & 3.79           & 2.48          & 0.20          & 0.67          & 1.47           \\
            Bilevel-Bleu                            & 21.16                             & \textbf{7.79}                       & \textbf{4.10 } & \textbf{2.71} & \textbf{0.22} & \textbf{0.70} & \textbf{1.61}  \\
            Bilevel-Cosine                          & \textbf{21.17}                    & 7.74                                & 4.10           & 2.68          & 0.21          & 0.64          & 1.34           \\
            \toprule
        \end{tabular}
        \vspace{-0.1cm}
        \label{tab:result}
    \end{table*}


    The experimental findings demonstrate a notable performance improvement of
    the proposed bilevel scheduled sampling model over both the traditional scheduled
    sampling model and the proposed model lacking sentence-level sampling,
    as evidenced by the higher BLEU-1/2/3/4 scores achieved.
    This superiority can be attributed to the incorporation of sentence-level information in the proposed model,
    which leads to the generation of more coherent and natural sentences,
    thereby achieving better alignment with the ground truth.
    Furthermore, it was observed that utilizing BLEU as the sentence-level score yielded better
    results compared to sentence-level cosine similarity.
    This discrepancy may arise from the fact that BLEU places greater emphasis on text matching,
    while cosine similarity focuses more on semantic alignment, thus favoring the former for improved BLEU results.

    Additionally, the proposed model shows an advantage in the Distinct-1/2/3 metric.
    This advantage stems from the incorporation of sentence-level sampling information
    and the utilization of probabilistic sampling for word generation.
    The bilevel model effectively generates sentences that are more diverse,
    mitigating the issue of excessively repetitive or singular output,
    and consequently attains higher scores in the Distinct metric.

    In summary, the experimental results demonstrate that by
    combining sentence-level and word-level
    considered together with probabilistic sampling, the proposed model outperforms
    existing scheduled sampling models in terms of both BLEU-1/2/3/4 and Distinct-1/2/3 metrics.
    This indicates that the bilevel scheduled sampling model significantly alleviates the exposure bias problem
    and outperforms the state-of-the-art scheduled sampling methods.

    \subsection{Human Evaluation}

    To thoroughly assess the proposed model and the baseline model mentioned in this paper,
    we conducted a human evaluation following the approach used by Li et al\cite{li2017adversarial}.
    For this evaluation, we randomly selected 100 samples from the test set of each dialogue dataset.
    Subsequently, we sought the judgment of three well-educated annotators to determine whether
    the overall response quality of the Bilevel-Bleu model and the other models under consideration
    exhibited superior coherence, informativeness, and fluency.
    The annotators categorized their assessment as either a win, tie, or lose for each model.

    \begin{table}[th]
        \centering
        \tabcolsep=0.3cm
        \caption{Human evaluation result.}

        \begin{tabular}{l c c c c}
            \toprule
            Bilevel-Bleu vs. Models                 & Win   & Tie   & Lose  & Kappa  \\  \midrule
            Transformer\upcite{2017Attention}       & 60.33 & 25.50 & 14.17 & 0.6581 \\
            AdapBridge\upcite{2021Adaptive}         & 48.50 & 29.67 & 21.83 & 0.5417 \\
            Confidence-Aware\upcite{2021Confidence} & 43.00 & 32.17 & 24.83 & 0.5077 \\
            Bilevel-None                            & 44.50 & 32.00 & 23.50 & 0.4491 \\
            Bilevel-Cosine                          & 39.33 & 33.33 & 27.33 & 0.5223 \\
            \bottomrule
        \end{tabular}
        \label{tab:human}
    \end{table}

    Table \ref{tab:human} summarizes the human evaluation results.
    The final results show that the Bilevel-Bleu model in this paper is better than other models,
    which indicates that it is more capable of generating human preferred responses.
    Meanwhile, we use Fleiss kappa \cite{fleisskappa/measuring} to measure the agreement between annotators,
    and the results are all greater than 0.4, which indicates that the annotators reach a good agreement on the judgment.

    \subsection{Ablation study}
    ​In this paper, we design a comparison experiment includes the proposed model without sentence-level
    sampling and with two different sentence-level sampling.
    As can be seen from Table \ref{tab:result}, whether the sentence-level
    indicator uses bleu or cosine similarity, the model combining
    sentence-level and word-level sampling in this paper will have better
    results than the single word-level sampling model.
    At the same time, it can also be seen from the results that the performance
    of the proposed model is significantly improved compared to the base
    transformer model when only word-level sampling is performed, with slightly
    better results than the existing scheduled sampling methods.
    This indicates that the probabilistic sampling approach in this paper
    is better than threshold truncation if we map the original probability to a suitable size by smoothing function.

    In addition, we conducted comparative experiments on smooth functions,
    including linear truncation function f1 (equation \ref{equation5}) and sigmoid-shaped smooth function f2 (equation \ref{equation6}).
    The result is in Table \ref{tab:result1}.

    \begin{table*}[!ht]
        \vspace{-0.1cm}
        \caption{Evaluation results of different smooth functions on the PersonaChat dataset.}

        \centering
        \begin{tabular}{l|cccc|ccc}
            \toprule
            Model                             & \multicolumn{4}{c|}{BLEU-1/2/3/4} & \multicolumn{3}{c}{Distinct-1/2/3 }                                                                                  \\
            \toprule
            Transformer\upcite{2017Attention} & 17.79                             & 6.37                                & 3.42           & 2.31          & 0.22          & 0.65          & 1.34          \\
            Bilevel-f1                        & 19.58                             & 7.18                                & 3.79           & 2.45          & 0.19          & 0.64          & 1.35          \\
            Bilevel-f2                        & \textbf{21.16}                    & \textbf{7.79}                       & \textbf{4.10 } & \textbf{2.71} & \textbf{0.22} & \textbf{0.70} & \textbf{1.61} \\
            \toprule
        \end{tabular}
        \label{tab:result1}
    \end{table*}

    The results indicate that  the sigmoid-shaped smooth function
    gives better results, enabling the model to sample more appropriate sentences.
    This is due to the direct probabilistic sampling  is not sufficient to distinguish
    the high-quality and low-quality utterances.
    By employing the sigmoid-shaped smooth function,
    the influence of sentence performance on the sampling probability is enhanced,
    resulting in a smoother and more effective sampling process.

    \section{Conclusion}
    In this paper, we propose a bilevel scheduled sampling model, which considers
    the sentence-level and word-level combination quality of the model generation results,
    so that the sampling results can better adapt to the exposure bias and thus improve the performance of the model.
    In order to make the sampling more diverse and improve the adaptability of model,
    we propose a smooth function to map the combined result of sentence-level
    and word-level to an appropriate range, and then perform probabilistic sampling instead of threshold truncation.
    ​Experiments on two widely used open-domain dialogue datasets demonstrate
    the effectiveness of all our proposed methods,
    which significantly alleviate the exposure bias problem
    and outperform state-of-the-art scheduled sampling methods.
    In the future, we plan to extend the application of the Bilevel Scheduled Sampling method to large language models across various projects, addressing the issue of exposure bias.
    This approach will help enhance the performance and robustness of the models in real-world scenarios.

    \section*{Acknowledgement}
    This research was supported by the Beijing Natural Science Foundation \\
    (No.4222037, L181010).

    \bibliographystyle{splncs04_unsort}


    \bibliography{ref}
\end{CJK}
\end{document}